\documentclass[letterpaper, 10 pt, conference]{ieeeconf}  

\IEEEoverridecommandlockouts                              

\title{\LARGE \bf
SelfTune: Metrically Scaled Monocular Depth Estimation through Self-Supervised Learning 
}

\author{Jaehoon Choi$^{*}$\:\textsuperscript{\rm 1,2}, Dongki Jung$^{*}$\:\textsuperscript{\rm 1},  Yonghan Lee$^{1}$, Deokhwa Kim$^{1}$, Dinesh Manocha$^{2}$, and Donghwan Lee$^{1}$\\
$^{1}$NAVER LABS $^{2}$University of Maryland
\thanks{* These two authors contributed equally}%
\thanks{Correspondence to kevchoi@umd.edu, dongki.jung@naverlabs.com}%
\thanks{This work was supported by the Institute of Information \& communications Technology Planning \& Evaluation(IITP) grant funded by the Korea government(MSIT) (No. 2019-0-01309, Development of AI Technology for Guidance of a Mobile Robot to its Goal with Uncertain Maps in Indoor/Outdoor Environments)}%
}

\usepackage{graphicx}
\usepackage{amsmath}
\usepackage{amssymb}
\usepackage{multirow}
\usepackage{booktabs}

\begin{document}
\maketitle
\thispagestyle{empty}
\pagestyle{empty}

\begin{abstract}
Monocular depth estimation in the wild inherently predicts depth up to an unknown scale. 
To resolve scale ambiguity issue, we present a learning algorithm that leverages monocular simultaneous localization and mapping (SLAM) with proprioceptive sensors.
Such monocular SLAM systems can provide metrically scaled camera poses.  
Given these metric poses and monocular sequences, we propose a self-supervised learning method for the pre-trained supervised monocular depth networks to enable metrically scaled depth estimation.
Our approach is based on a teacher-student formulation which guides our network to predict high-quality depths. 
We demonstrate that our approach is useful for various applications such as mobile robot navigation and is applicable to diverse environments. 
Our full system shows improvements over recent self-supervised depth estimation and completion methods on EuRoC, OpenLORIS, and ScanNet datasets.      
\end{abstract}

\section{Introduction}

Estimations of structure and ego-motion from a monocular camera supports a variety of tasks, from AR/VR to robot navigation.
The conventional approach to handling these tasks is monocular simultaneous localization and mapping (SLAM) \cite{orb-slam}. 
Due to the nature of a single moving camera, estimated camera ego-motion and structure are only defined \textit{up to scale}.
In practical applications, monocular SLAM systems are combined with data from proprioceptive sensors such as an inertial measurement unit (IMU) \cite{msckf,VINS-Mono} and wheel odometry measurements \cite{VINSonWheels,VIWO} to acquire the accurate metric scale. Based on these sensors, recent SLAM systems \cite{VINS-Mono,VINSonWheels} can provide accurate ego-motion estimates. Nevertheless, dense depth estimation is still an important issue because the sparsity of SLAM maps means that there is not enough information for obstacle avoidance and path planning. Depth completion from the sparse depth maps paired with images is an appealing method to complement SLAM mapping \cite{Aerial}--\cite{VOID}. However, these depth completion methods are vulnerable to noisy depth values from SLAM and varying sparse point distributions \cite{Aerial,Robust-Mono-VI}. Thus, we aim to leverage the learning-based depth estimation from a single image to predict depth with a metric scale \cite{towardiros}--\cite{DnD}.   

Recently, many data-driven approaches for depth estimation have shown promising results by training neural networks on large-scale datasets with dense groundtruth depth maps \cite{DORN,Adabins}. Training data with a dense metric depth is difficult to collect because it requires a significant amount of time and expensive hardware with range sensors (e.g., LiDAR) to collect. Instead, many researchers explore the idea of collecting diverse in-the-wild data such as internet photo collections and web stereo videos \cite{Megadepth}--\cite{LeReS} because the internet has a nearly unlimited data source. Due to the availability of large-scale datasets and training strategies, recent monocular depth estimation methods \cite{Midas,LeReS} have achieved both high-resolution depth estimation and generalizability across diverse environments. However, as metric depth is not available on diverse datasets, their models are not able to estimate the metric depth and reconstruct an accurate 3D scene structure. 
This is why we focused on a self-supervised fine-tuning method which allows the supervised models \cite{Midas,LeReS} to adapt to unknown environments by achieving metric monocular depth estimation.

\begin{figure}[t]
\centering
\includegraphics[width=1.0\linewidth]{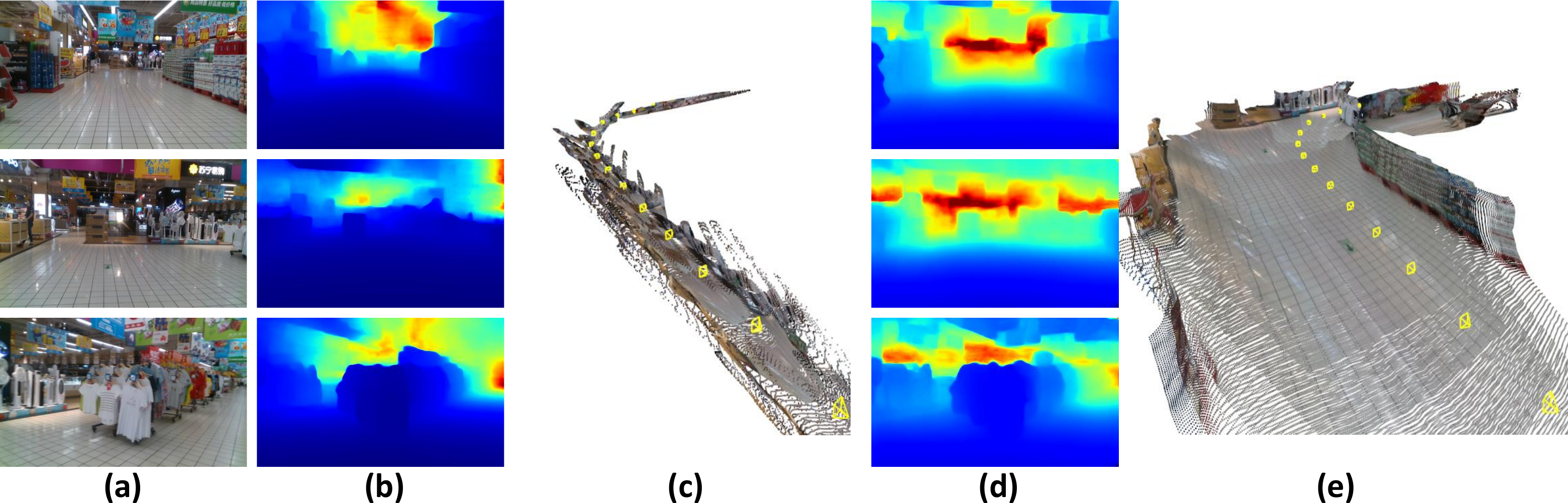}
\vspace{-6mm}
\caption{(a) shows the input RGB images. Both (b) MiDaS \cite{Midas} and (d) our method have good monocular depth estimation results. 
However, (e) our method can recover the accurate 3D scene with metric scale, while 3D point clouds from (c) MiDaS suffers from distortion due to scale ambiguity. 
}
\label{figure1}
\vspace{-5mm}
\end{figure}

\noindent\textbf{Main Results:} 
We present a novel approach for fine-tuning the pre-trained monocular depth network \cite{Midas,LeReS}, which only generates depth up to an unknown scale. Since these supervised monocular depth networks leverage high-level scene priors from data, our network should estimate metric depth without forgetting a strong data-driven prior. 
Inspired by knowledge distillation \cite{KD}, we form a teacher-student learning that constrains the student network to preserve the ability to estimate detailed depth maps. In addition, our approach can adopt a lightweight network to run our dense mapping process in parallel to the SLAM process. Given the metric-scaled poses from SLAM and estimated depth, our proposed approach allows our depth network to learn metrically accurate and consistent depth estimation. We show that our method can be used along with either visual-inertial systems or visual SLAM incorporating wheel odometers.


\begin{itemize}
    \item We develop a new approach that fine-tunes the pre-trained monocular depth networks to estimate metric depth. To do this, we utilize a self-supervised method that does not require dense depth groundtruth. 
    \item We adopt a teacher-student learning that enables the student network to jointly learn the ability to synthesize detailed depth and estimate scale-consistent depths.
    \item Due to the incorrect depth points with varying sparsity produced by monocular SLAM systems, we demonstrate that an image-only depth network along with accurate pose estimates can be an alternative to the prevalent depth completion method. 
    We observe 9.6\%-29.5\% improvement in RMSE on various datasets.    
\end{itemize}

\section{Related Work}

\noindent\textbf{Depth Estimation:} The main properties driving monocular depth estimation are what data can be used for training. When dense depth maps collected from depth sensors are used for training, most methods can be trained to directly regress depth maps \cite{DORN,Adabins}.
To overcome the burden of supervision, self-supervised learning using monocular video sequences \cite{SfMLearner}--\cite{sc-sfmlearner} is a popular approach. 
This method exploits the estimated depths and relative poses such that a photometric constraint is used to serve as a self-supervision signal.
Other recent works present the approach of learning with diverse datasets scraped from stereo images or videos from the internet \cite{web3dv}--\cite{LeReS}. As the metric depth is not available for the diverse datasets, these methods use scale and shift-invariant losses for supervision. Although they generalize very well to unseen data, the 3D reconstructed scene shape can be distorted because of the unknown scale. A limitation of both self-supervised learning and learning with diverse datasets is that they can only produce scale-ambiguous depth predictions. To resolve this, Guizilini et al. \cite{packnet-sfm} use the camera's velocity from an additional sensor. Additionally, some works \cite{Casser,cvd} present a test-time fine-tuning strategy for predicting consistent depth. To combine the advantages of these works, we propose a fine-tuning method for estimating metrically accurate and consistent depth from a monocular sequence.    

\noindent\textbf{Depth Completion:} The depth completion reconstructs dense depth maps from sparse depth maps paired with single images. Prior works have focused on this task because SLAM often acquires sparse depth maps from either a low-resolution depth sensor \cite{Volumetric} or sparse features of SLAM algorithms \cite{orb-slam,VINS-Mono}. Most depth completion methods rely on supervised learning, requiring large numbers of dense depth maps collected from depth sensors \cite{2018ICRAMa}--\cite{Volumetric}. Recent works \cite{VOID}--\cite{SelfDeco} explore self-supervised learning from monocular video sequences paired with sparse depth maps. In particular, the coupling of visual-inertial odometry (VIO) and depth estimation \cite{Aerial}--\cite{Robust-Mono-VI} has attracted a lot of attention. Some methods \cite{VOID,Kimera} utilize a mesh of sparse points to predict dense depth maps. \cite{Aerial,Robust-Mono-VI} adopt a similar method with supervised depth completion to improve the robustness. Sartipi et al. \cite{DD-VISLAM} leverage detected planes and surface normals to complement sparse depths from VIO. 
Except for \cite{VOID}, such methods are based on supervised learning. 
Furthermore, these depth completion methods designed for VIO suffer from noisy, sparse depth values and varying sparse points.  


\section{Our Approach: SelfTune}   
\subsection{SLAM systems}
Due to the nature of monocular SLAM, additional information is required to recover the metric scale. In this paper, we demonstrate that two different SLAM systems based on either IMUs or wheel odometry can be incorporated into our method. The state-of-the-art visual-inertial SLAM systems \cite{msckf}--\cite{VIOBenchmark} have achieved accurate motions, even on low-cost mobile devices, because IMU can solve the scale factor ambiguity with monocular cameras. We adopt VINS-Mono \cite{VINS-Mono}, which is based on a tightly-coupled sliding window optimization of pre-integrated IMU and visual measurements. We only exploit front-end modules to estimate odometry without using relocalization and loop-closure. Moreover, in the case of wheeled robots, SLAM systems can be used in conjunction with wheel odometry measurements \cite{VINSonWheels,VIWO}. We implement a visual-wheel odometry algorithm that uses sliding window-based graph optimization to estimate metrically accurate robot motions.  

\subsection{Training Method}

\begin{figure*}[t]
\centering
\includegraphics[width=0.85\linewidth]{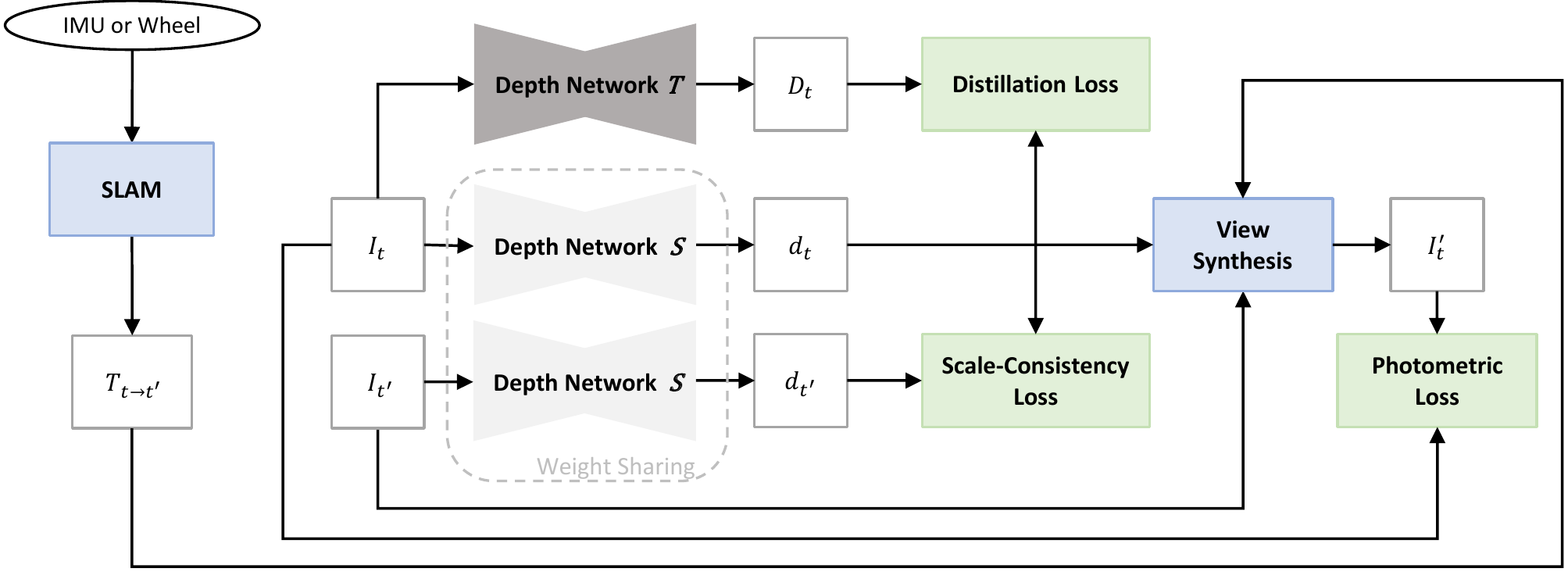}
\caption{Overview of our training pipeline.
The monocular SLAM systems using low-cost sensors like IMU or wheel-odometry can provide the metric scale of relative camera position \(T_{t \rightarrow t^{\prime}}\).
Depth networks consist of student \(S\) and teacher \(T\). However, only network \(S\) is trained and applied for inference.
Our learning method combines three terms.
1) The distillation loss encourages our network to preserve the detailed shape of the depth map from the supervised depth network.
2) The photometric loss, which is based on the view synthesis, is the core function for delivering absolute scale information from \(T_{t \rightarrow t^{\prime}}\).
3) The scale-consistency loss enforces scale consistency in frame-to-frame depth maps.
}
\label{figure2}
\vspace{-6mm}
\end{figure*}

In Fig. \ref{figure2}, let \(T\) and \(S\) be a teacher network and a student network, respectively. Our method employs pre-trained models both for the teacher and the student networks. We fix the weights of the teacher network to guide the learning of the student network. Both teacher and student networks take a target view image \(I_{t}\) as input and outputs inverse depth maps \(D_{t}\) and \(d_{t}\). The student network also takes source view images containing its two adjacent temporal frames, i.e., \(I_{t^{\prime}} \in \{I_{t-1}, I_{t+1}\}\), although including a larger temporal context is possible. The SLAM system can provide relative poses \(T_{t\rightarrow t^{\prime}}\) between the target image \(I_{t}\) and the set of source images \(I_{t^{\prime}}\) by using scale information from proprioceptive sensors. Our model can work in parallel with the SLAM system, requiring the estimated camera poses.   

\subsubsection{Photometric Loss} \label{Photometric Loss}
Our method aims to estimate a metrically scaled depth of the selected keyframes. The pre-trained depth estimation network estimates depth maps that are arbitrarily scaled and have large magnitudes. We apply Min-Max normalization to the inverse depth map \(d_{t}\):
\begin{equation}
\widetilde{d}_{t} = \frac{d_{t} - \text{min}(d_{t})}{\text{max}(d_{t}) - \text{min}(d_{t})}
\label{normalized_depth}
\end{equation}
Given the camera intrinsic matrix \(K\), we utilize the inverse depth \(\widetilde{d}_{t}\) and the relative camera pose \(T_{t\rightarrow t^{\prime}}\) to synthesize target image  \(I^{\prime}_{t}\) by using pixels from neighboring source images \(I_{t^{\prime}}\). 
\begin{equation}
\begin{split}
    I^\prime_{t} &= I_{t^{\prime}}\langle \pi( K, T_{t\rightarrow t^{\prime}}, \rho(\widetilde{d}_{t})) \rangle\:.
\end{split}
\label{image_warping}
\end{equation}
where \(\rho(\cdot)\) converts the normalized inverse depth into the depth \cite{Monodepth2}, \(\pi(\cdot)\) returns pixel coordinates of \(d_{t}\),  and \(\langle \cdot \rangle\) is the bilinear sampling operator \cite{STN}. We set the photometric loss \(L_{p}\) as the combination of the \(L_{1}\) and SSIM \cite{SSIM}. Following \cite{Monodepth2}, we compute the minimum photometric loss per pixel for each source image to select the best matching pixels. In addition, we apply an auto-mask in \cite{Monodepth2} to remove pixels whose appearance does not change between neighboring frames as a result of the stationary camera or objects with no relative motion:   
\begin{equation}
\begin{split}
    L_{p}(I_{t}, I^\prime_{t}) &= \frac{\alpha}{2} (1-\text{SSIM}(I_{t}, I^\prime_{t})) 
    + (1-\alpha)L_{1}(I_{t},I^\prime_{t}),\\ 
    & \qquad L_{photo} = \min_{I_{t^\prime}} L_{p}(I_{t}, I^\prime_{t}).
\end{split}
\label{photometric_loss}
\end{equation}
where \(\alpha = 0.85\) in all experiments. Since this relative camera pose is metrically accurate, our depth network can learn metrically accurate depth estimates in order to maintain consistency. 

\subsubsection{Distillation Loss} 
At training time, we fine-tune a supervised depth network with self-supervision based on the photometric loss. Thus, the teacher network prevents the student from being degraded due to the discrepancy between the different learning methods.
As in Fig. \ref{figure2}, the scale ambiguous inverse depth map \(D_{t}\) regularizes the predicted inverse depth map \(d_{t}\) to conserve the relative scale order and the boundary detail of the depth map. 
We utilize scale- and shift-invariant loss \cite{Midas}, which align the distribution of relative depth, to maintain the consistent structure in the depth map:
\begin{equation}
L_{ssi} = \frac{1}{2N}\sum_{j=1}^{N^{\prime}}\left|\eta(\widetilde{d}_{j})-\eta(\widetilde{D}_{j})\right|,
\label{scale_and_shift_invariant_loss}
\end{equation}
where \(N^{\prime} = 0.8N\) for sorting and trimming the largest \(20\%\) of the residuals, following \cite{Midas}. The align function \(\eta(\cdot)\) normalizes the depth with mean and median values to have no shift and unit scale:
\begin{equation}
\eta({D}) = \frac{D-\text{median}(D)}{\text{mean}(\left|D-\text{median}(D)\right|)}.
\label{get_scale_and_shift_val}
\end{equation}
Furthermore, to encourage a smoother edge-aware depth map, a scale-invariant gradient matching term \cite{Megadepth} is used,
\begin{equation}
\begin{split}
L_{reg} = \frac{1}{N}\sum_{i=1}^{N}(&|\triangledown_{x}(\eta(\widetilde{d}_{i})-\eta(\widetilde{D}_{i}))| \\
+&|\triangledown_{y}(\eta(\widetilde{d_{i}})-\eta(\widetilde{D}_{i}))|). \\ 
\end{split}
\label{gradient_ssil}
\end{equation}
The total loss function of distillation is described as follows:
\begin{equation}
    L_{d} = L_{ssi} + \frac{1}{2} L_{reg}.
\end{equation}

\subsubsection{Scale-Consistency Loss} 
Inspired by \cite{sc-sfmlearner}, given accurate and metrically scaled motions, we can enforce the scale consistency by minimizing the geometric distance of estimated depths between neighboring frames. Our method computes the depth maps \(d_{t^{\prime}}\) of each temporally neighboring frame \(I_{t^{\prime}} \in \{I_{t-1}, I_{t+1}\}\). Then we are able to generate \(\hat{d}_{t^{\prime}}\) by warping temporally adjacent depth maps \(d_{t^{\prime}}\) using the inverse depth \(\widetilde{d}_{t}\) and the relative camera pose \(T_{t\rightarrow t^{\prime}}\). Additionally, we obtain the \(d_{t \rightarrow t^{\prime}}\) by lifting \(d_{t}\) to 3D space and warping \(d_{t}\) using \(T_{t\rightarrow t^{\prime}}\). We utilize the L1-norm to penalize the difference between \(\hat{d}_{t^{\prime}}\) and \(d_{t\rightarrow t^{\prime}}\)   
\begin{equation}
\begin{split}
    \hat{d}_{t^{\prime}}& = d_{t^{\prime}}\langle \pi( K, T_{t\rightarrow t^{\prime}}, \rho(\widetilde{d}_{t})) \rangle, \\
    L_{c} &= \sum_{x \in \Omega} |d_{t \rightarrow t^{\prime}}(x) - \hat{d}_{t^{\prime}}(x)|.
\end{split}
\label{geometric_consistency}
\end{equation}
where \(\Omega\) denotes the valid regions that are successfully matched from \(I_{t^{\prime}}\) to the \(I_{t}\). To obtain the valid regions \(\Omega\), we exclude the regions that include auto-masks obtained from Sec. \ref{Photometric Loss}. Instead of calculating minimum photometric loss, reprojected error norm is averaged over multiple depth maps predicted from temporally adjacent frames. This depth consistency loss discourages scale drift between dense depth predictions in adjacent frames and is helpful for producing a tight alignment with ego-motion predictions.


\subsubsection{Final Loss Function}
As in \cite{SfMLearner,Monodepth2}, we also add an edge-aware smoothness loss \(L_{smooth}\) to propagate depth values from discriminative regions to textureless regions. We combine all loss functions into a single loss for training:
\begin{equation}
L_{final} = L_{photo} + \omega_{d}L_{d} + \omega_{sm}L_{smooth} + \omega_{c}L_{c}
\label{total_loss}
\end{equation}
where \(\omega_{d}\), \(\omega_{sm}\), and \(\omega_{c}\) are weights for different losses selected through a grid search.

\subsubsection{Lightweight backbone}
In practice, our approach aims to perform dense depth estimation and SLAM systems simultaneously. However, most pre-trained depth estimation models consist of a large backbone (e.g., ResNet50 \cite{ResNet} or ResNeXt101 \cite{ResNeXt}), which requires a high computational overhead. To overcome this burden, we utilize a lightweight pre-trained model (based on EfficientNet-Lite \cite{EfficientNet})\footnote{is available at https://github.com/isl-org/MiDaS/tree/master/mobile} as a student network while the pre-trained model of the large backbone is used as a teacher network. Low-resolution images are fed to the student network, while high-resolution images are input to the teacher network. This is because the network shows better performance when the input image's size is close to the training resolution.  To match the resolutions of both networks' depth outputs, we upsample the lower resolution depth maps predicted from the student network.            


\section{Implementation and Performance}
To validate our method, we use the standard metrics for depth estimation \cite{Eigen}: absolute relative error (Abs Rel), square relative error (Sq Rel), root mean square error (RMSE) and its log scale (RMSE log), and inlier ratios (\(\delta_{1.25}\), \(\delta_{1.25^2}\), and \(\delta_{1.25^3}\)). We utilize EuRoC \cite{EuRoC} and OpenLORIS \cite{OpenLORIS} to evaluate visual-inertial data. We show qualitative results on the ADVIO dataset \cite{ADVIO} collected by a handheld device. The lack of visual-wheel odometry data led us to employ our wheeled robot to collect visual and wheel measurements.      

\subsection{Datasets} 
\subsubsection{EuRoC MAV Dataset}
The EuRoC MAV dataset \cite{EuRoC} is collected onboard a micro-aerial vehicle (MAV), and it contains stereo images, synchronized IMU measurements, and 3D scans on Vicon Room. We only use the monocular video from the left camera and IMU measurements. Since this camera has significant lens distortion, we remove this distortion for training. We train on the V1\_01\_easy and V1\_02\_medium and evaluate V2\_01\_easy with 3D groundtruth depth maps.           

\subsubsection{OpenLORIS Dataset}
The OpenLORIS dataset \cite{OpenLORIS} is captured by a wheeled robot equipped with a RealSense D435i camera, a RealSens T265 camera and a synchronized IMU sensor. This dataset consists of five sequences captured from diverse environments. To validate our method, we generate four different experimental settings: OL-Market (cafe and market), OL-Home (home), OL-Corridor (corridor), and OL-Office (office). We train on each of four different environments and test on market-1, home-2, corridor-2, and office-2 sequences.      

\subsubsection{ScanNet dataset} 
ScanNet \cite{ScanNet} is a dataset composed of RGB-D frames from indoor environments. The depths and camera poses are obtained from RGB-D 3D reconstruction. We use the official training split and downsampled 10 times for a toy experiment. Since ScanNet does not contain proprioceptive sensor data, we use groundtruth camera poses and 100 depth values sampled from depth labels.  

\subsubsection{ADVIO dataset}
The ADVIO dataset \cite{ADVIO} comprises scenes collected from smartphone sensors. Although the ADVIO dataset does not provide dense 3D scans provided by depth sensors, it includes diverse real-world scenes that other datasets do not provide. 

\subsection{Implementation Details}
We run VINS-Mono \cite{VINS-Mono} in visual-inertial mode on EuRoC, OpenLORIS, and ADVIO to estimate camera poses per frame. 
We reimplement a visual-wheel odometry algorithm and collect data from our office. 
The proposed learning-based approach has been implemented using the PyTorch library \cite{PyTorch}. Our models are trained for 15 epochs using Adam optimizer \cite{Adam}, with a batch size of 8 and a learning rate of \(10^{-4}\). We use an image resolution of 256x384 for EuRoC and OpenLORIS. In the final loss function, we set the \(\omega_d\) to 0.001,  \(\omega_{sm}\) to 0.01, and \(\omega_c\) to 0.01. 

\begin{table}[t]
    \begin{center}
    \caption{\label{table:metric} Quantitative comparisons with recent algorithms}
    \resizebox{0.49\textwidth}{!}
    {
    \begin{tabular}{l|c|cccc|ccc}
    \toprule
    \multirow{2}{*}{Method} & \multirow{2}{*}{Data} & \multicolumn{4}{c|}{Lower is better} & \multicolumn{3}{c}{Higher is better} \\
    & &  Abs Rel & Sq Rel & RMSE & RMSE log & $\delta_{1.25}$ & $\delta_{1.25^2}$ & $\delta_{1.25^3}$ \\     
    \midrule
    MiDaS \cite{Midas} & EuRoC & 0.845 & 0.729 & 1.889 & 2.096 & 0.008 & 0.017 & 0.029 \\
    Monodepth2 \cite{Monodepth2} & EuRoC & 0.441 & 0.430 & 1.039 & 0.405 & 0.405 & 0.714 & 0.870 \\
    SelfDeco \cite{SelfDeco} & EuRoC & 0.401 & 0.370 & 0.958 & 0.394 & 0.455 & 0.766 & 0.902 \\
    Ours & EuRoC & \textbf{0.286} & \textbf{0.200} & \textbf{0.675} & \textbf{0.317} & \textbf{0.603} & \textbf{0.856} & \textbf{0.946} \\
    \midrule
    MiDaS \cite{Midas} & OL-Market & 0.512 & 2.281 & 4.716 & 0.488 & 0.583 & 0.810 & 0.885 \\
    Monodepth2 \cite{Monodepth2} & OL-Market & 0.278 & 0.838 & 1.387 & 0.337 & 0.717 & 0.904 & 0.955 \\
    SelfDeco \cite{SelfDeco} & OL-Market & 0.313 & 0.660 & 1.492 & 0.357 & 0.680 & 0.872 & 0.931 \\
    Ours & OL-Market & \textbf{0.231} & \textbf{0.420} & \textbf{1.254} & \textbf{0.296} & \textbf{0.742} & \textbf{0.928} & \textbf{0.964} \\
    \midrule
    MiDaS \cite{Midas} & OL-Home & 0.715 & 2.652 & 4.158 & 0.612 & 0.418 & 0.659 & 0.785 \\
    Monodepth2 \cite{Monodepth2} & OL-Home & 0.289 & 0.266 & 0.699 & 0.316 & 0.573 & 0.850 & 0.947 \\
    SelfDeco \cite{SelfDeco} & OL-Home & 0.342 & 0.400 & 0.804 & 0.399 & 0.501 & 0.781 & 0.900 \\
    Ours & OL-Home & \textbf{0.189} & \textbf{0.075} & \textbf{0.455} & \textbf{0.228} & \textbf{0.697} & \textbf{0.948} & \textbf{0.987} \\
    \midrule
    MiDaS \cite{Midas} & OL-Corridor & 0.542 & 2.079 & 6.255 & 0.493 & 0.526 & 0.761 & 0.862 \\
    Monodepth2 \cite{Monodepth2} & OL-Corridor & 0.244 & 0.174 & 1.502 & 0.356 & 0.690 & 0.848 & 0.908 \\
    SelfDeco \cite{SelfDeco} & OL-Corridor & 0.346 & 0.240 & 1.789 & 0.262 & 0.468 & 0.736 & 0.851 \\
    Ours & OL-Corridor & \textbf{0.164} & \textbf{0.079} & \textbf{0.929} & \textbf{0.185} & \textbf{0.815} & \textbf{0.959} & \textbf{0.983} \\
    \midrule
    MiDaS \cite{Midas} & OL-Office & 0.845 & 2.170 & 3.377 & 0.726 & 0.238 & 0.468 & 0.653 \\
    Monodepth2 \cite{Monodepth2} & OL-Office & 0.244 & 0.203 & 0.941 & 0.298 & 0.690 & 0.894 & 0.953 \\
    SelfDeco \cite{SelfDeco} & OL-Office & 0.192 & 0.080 & 0.495 & 0.245 & 0.712 & 0.919 & 0.979 \\
    Ours & OL-Office & \textbf{0.105} & \textbf{0.028} & \textbf{0.300} & \textbf{0.156} & \textbf{0.913} & \textbf{0.979} & \textbf{0.993} \\
    \midrule
    MiDaS \cite{Midas} & ScanNet & 0.397 & 0.474 & 1.621 & 0.421 & 0.486 & 0.761 & 0.883 \\
    Monodepth2 \cite{Monodepth2} & ScanNet & 0.145 & 0.044 & 0.472 & 0.202 & 0.758 & 0.902 & 0.971 \\
    SelfDeco \cite{SelfDeco} & ScanNet & \textbf{0.103} & \textbf{0.030} & \textbf{0.391} & \textbf{0.152} & \textbf{0.855} & \textbf{0.958} & \textbf{0.989} \\
    Ours & ScanNet & 0.116 & \textbf{0.030} & 0.401 & 0.162 & 0.778 & 0.940 & 0.988 \\
    \bottomrule
    \end{tabular}
    }
    \end{center}
    A median scaling is applied for MiDaS \cite{Midas} to choose a scale factor. Our method shows improvement in RMSE metric by 29.5\% in EuRoC and by 9.6\% in OL-Market. 
\vspace{-5mm}
\end{table}

\begin{figure*}[t]
\centering
\includegraphics[width=0.82\linewidth]{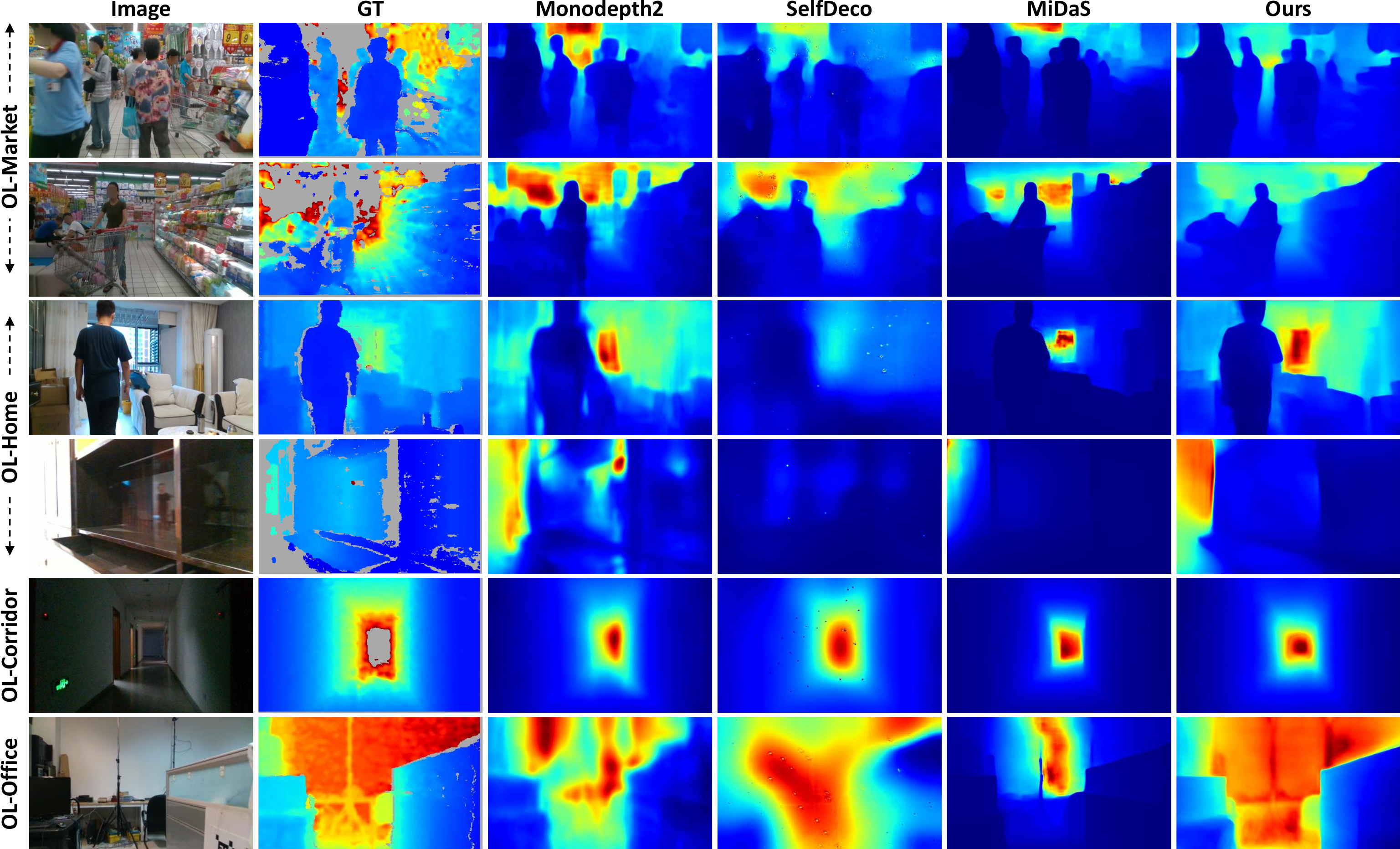}
\caption{Qualitative comparison with Monodepth2 \cite{Monodepth2}, SelfDeco \cite{SelfDeco}, and MiDaS \cite{Midas} on the OpenLORIS Dataset. Our model takes an image (1st column) and generates dense depth maps (6th column). Compared to depth results (3-5 columns) estimated by the previous depth estimation algorithms, our method predicts detailed and high quality depth maps. (low \protect\includegraphics[width=1cm,height=0.2cm]{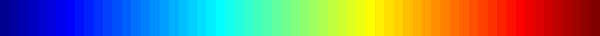} high; grey means empty depth values.)
}
\label{figure:depth_visual}
\vspace{-3mm}
\end{figure*}

\subsection{Evaluation}
\noindent\textbf{Quantitative Comparison:} We compare our method to other baseline approaches in Table \ref{table:metric}. MiDaS \cite{Midas} is the state-of-the-art method for supervised monocular depth estimation, which has a strong generalization capability on diverse datasets. We use its released checkpoints for evaluation and then apply a median scaling to determine a scale factor by comparing our estimated depths with the groundtruth depth maps. MonoDepth2 \cite{Monodepth2} is a benchmark algorithm for self-supervised monocular depth estimation. SelfDeco \cite{SelfDeco} is the state-of-the-art method for self-supervised monocular depth completion. For a fair comparison, we retrain MonoDepth2 and SelfDeco on our datasets. Also, instead of their pose estimation methods, poses generated from either Visual-Inertial Odometry (VIO) or Visual-Wheel Odometry (VWO) are exploited for training these models because these poses are more metrically accurate. Thus, they can make the depth prediction metrically consistent by applying photometric loss. SelfDeco uses sparse depth maps obtained from either VIO or VWO and corresponding images for training.

\begin{table}[t]
    \begin{center}
    \caption{\label{ablation_performance} Ablation study for the proposed model.}
    \resizebox{0.49\textwidth}{!}{
    \normalsize\begin{tabular}{l|c|ccc|cccc|ccc}
    \toprule
    \multirow{2}{*}{Model} & \multirow{2}{*}{Data} & \multirow{2}{*}{$L_{photo}$} & \multirow{2}{*}{$L_{d}$} & \multirow{2}{*}{$L_{c}$}&  \multicolumn{4}{c|}{Lower is better.} & \multicolumn{3}{c}{Higher is better.} \\ 
    & & & & & Abs Rel & Sq Rel\: & RMSE & RMSE log \: &\:  $\delta<1.25$ & $\delta<1.25^2$ & $\delta<1.25^3$ \\ 
    \midrule
    Ours & EuRoC & \checkmark & & & 0.393 & 0.400 & 0.946 & 0.555 & 0.541 & 0.768 & 0.863 \\
    Ours & EuRoC & \checkmark & \checkmark  & & 0.348 & 0.270 & 0.837 & 0.361 & 0.516 & 0.788 & 0.920 \\
    Ours & EuRoC & \checkmark & \checkmark  & \checkmark & \textbf{0.286} & \textbf{0.200} & \textbf{0.675} & \textbf{0.317} & \textbf{0.603} & \textbf{0.856} & \textbf{0.946} \\
    Ours* & EuRoC & \checkmark & \checkmark  & \checkmark & 0.327 & 0.230 & 0.756 & 0.352 & 0.519 & 0.835 & 0.936\\
    \midrule
    Ours & OL-Market & \checkmark & & & 0.238 & 0.470 & 1.307 & 0.298 & 0.753 & 0.920 & 0.962 \\
    Ours & OL-Market & \checkmark & \checkmark  & & 0.233 & 0.470 & 1.286 & 0.297 & \textbf{0.765} & 0.920 & 0.961 \\
    Ours & OL-Market & \checkmark & \checkmark  & \checkmark & \textbf{0.231} & \textbf{0.420} & \textbf{1.254} & \textbf{0.296} & 0.742 & \textbf{0.928} & \textbf{0.964} \\
    Ours* & OL-Market & \checkmark & \checkmark  & \checkmark & 0.270 & 0.517 & 1.358 & 0.326 & 0.711 & 0.899 & 0.947 \\
    \midrule
    Ours & ScanNet & \checkmark & & & 0.131 & 0.040 & 0.441 & 0.181 & 0.756 & 0.925 & 0.982 \\
    Ours & ScanNet & \checkmark & \checkmark  & & 0.127 & 0.040 & 0.433 & 0.175 & 0.768 & 0.936 & 0.986 \\
    Ours & ScanNet & \checkmark & \checkmark  & \checkmark & \textbf{0.116} & \textbf{0.030} & \textbf{0.401} & \textbf{0.162} & \textbf{0.778} & \textbf{0.940} & \textbf{0.988} \\
    Ours* & ScanNet & \checkmark & \checkmark  & \checkmark & 0.154 & 0.045 & 0.531 & 0.210 & 0.661 & 0.905 & 0.981 \\
    \bottomrule
    \end{tabular}
    }
    \end{center}
    Ours* indicates that we train the student network from scratch. Compared to the baseline (\(L_{photo}\) only), we improve the RMSE metric by 28.6\% in EuRoC, by 4.1\% in OL-Market, and by 9.1\% in ScanNet.   
\vspace{-5mm}
\end{table}

From the results on EuRoC and OpenLORIS datasets in Table \ref{table:metric}, our approach outperforms other methods. Since SelfDeco exploits accurate sparse depth valuse sampled from groundtruth depth maps in ScanNet, our method performs slightly worse than SelfDeco. Except for depth completion methods, our method performs better than image-only depth networks. This indicates that depth completion methods require high-quality sparse depth points provided by depth sensors for accurate depth estimation. However, current VIO and VWO systems can only generate noisy sparse depths; therefore, these sparse depths hurt the accuracy and completeness of completed depth maps.       


\noindent\textbf{Qualitative Comparison:} In Fig. \ref{figure:depth_visual}, we show the qualitative comparison of different depth estimation methods. 
We notice a significant improvement compared to Monodepth2 and SelfDeco in the visualization. As the fourth row in Fig. \ref{figure:depth_visual}, our method is robust to certain environments with heavy lights or reflections. Compared to MiDaS, our approach can slightly improve the qualitative results. These results are meaningful when considering the difference in sizes of backbone networks between the MiDaS (ResNeXt101 \cite{ResNeXt}) and ours (EfficientNet-Lite \cite{EfficientNet}). 
%

\begin{figure}[t]
    \centering
    \includegraphics[width=0.87\linewidth]{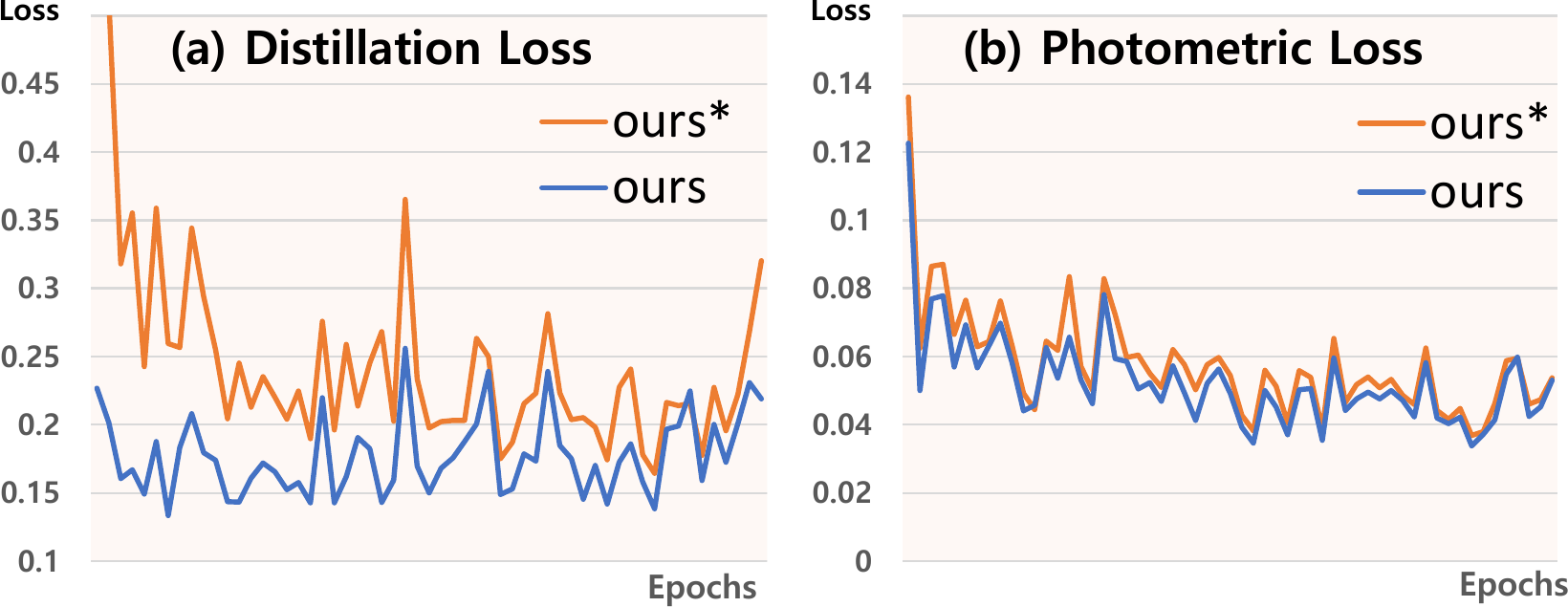}
    \caption{The plot of (a) distillation loss and (b) photometric loss during training on OL-Market. Ours* indicates that we train the student network from scratch.}
    \vspace{-6mm} 
    \label{figure:loss}
\end{figure}

\begin{figure*}[t]
\centering
\includegraphics[width=0.75\linewidth]{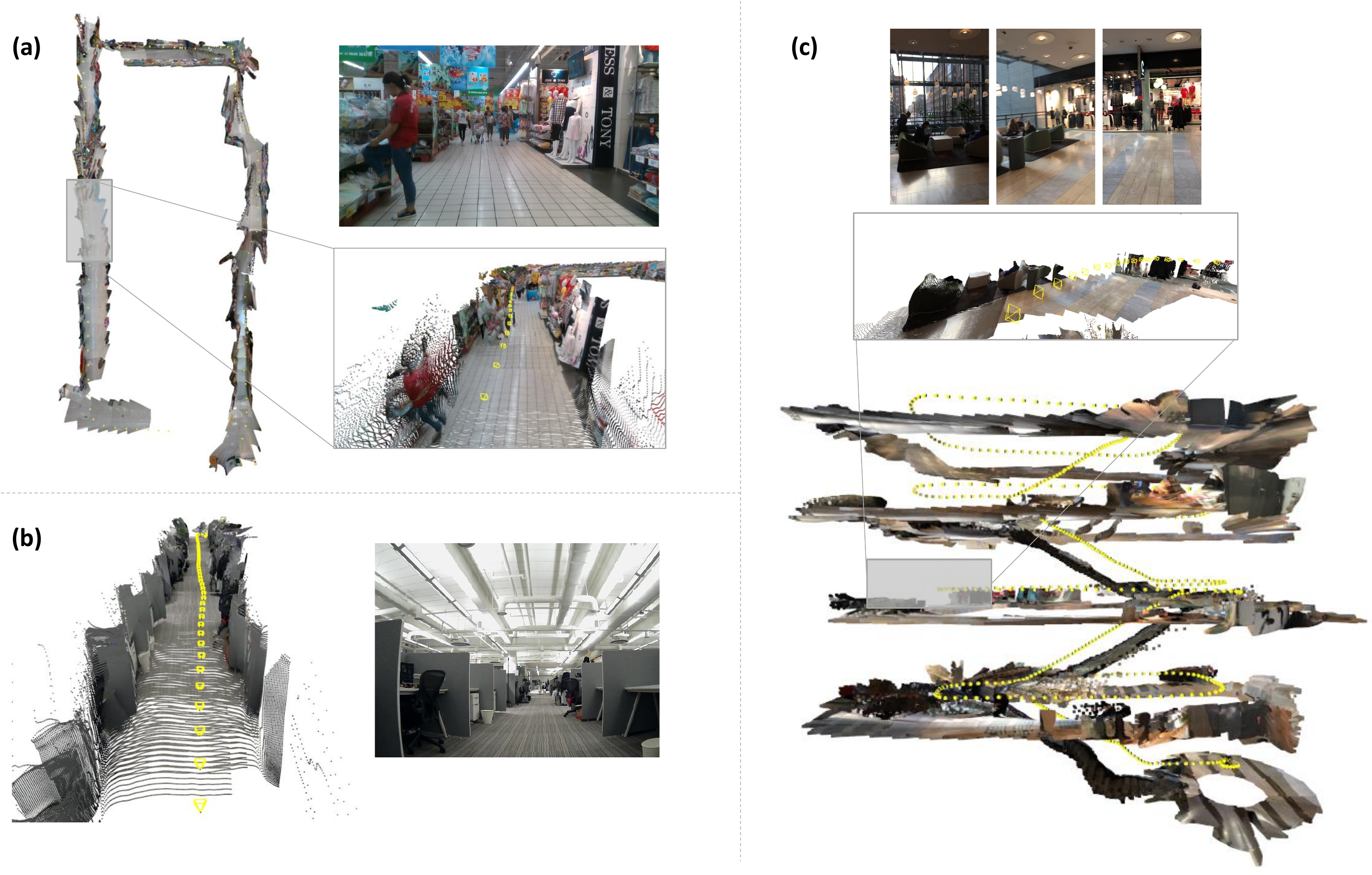}
\caption{Qualitative results: (a) market in the OpenLORIS dataset (captured by a wheeled robot), (b) office environment collected by our wheeled robot, and (c) shopping mall in the ADVIO dataset (collected from a handheld device). We show the zoomed-in regions and global views for details. We use wheel odometry measurements instead of IMU sensors to generate robot poses in (b). We accumulate the 3D point clouds by reprojecting the estimated depths from temporally consecutive scenes. Our method can generate globally coherent depth estimation aligned with camera poses.}
\vspace{-4mm} 
\label{figure:3D}
\end{figure*}

\noindent\textbf{Ablation Study:} 
In Table \ref{ablation_performance}, all of our proposed contributions improve the depth estimation performance over the baseline method. We also investigate the effects that pre-trained models have on our proposed training methodology. Although we fix the pre-trained models for the teacher network to maintain consistency with previous experiments, the student network is trained from scratch (Ours* in Table \ref{ablation_performance}). Our fine-tuning method outperforms training from scratch. In Fig. \ref{figure:loss}, we see that the distillation loss is a key difference between the two methods, while photometric loss shows similar trends. The distillation loss allows the learning of better depth estimation. This is because initializing the model's weights near the optimal solution enables fast convergence and may avoid the risk of convergence to local minima.        

\noindent\textbf{Different Pre-trained Network:} Our work is based on MiDaS \cite{Midas}, the most popular depth estimation method. Here, we investigate the depth estimation performance of our approach using LeReS \cite{LeReS}, which is a different pre-trained depth estimation network. This depth estimation network is designed to output depth maps instead of inverse depth maps like previous works \cite{Midas, SGRNet}. Thus, when applying our training method to fine-tune LeReS, we cannot employ Min-Max normalization to the inverse depth map. Similar to MiDaS, LeReS can only estimate depth up to an unknown scale. For comparison, a median scaling is applied to define a scale factor using groundtruth depth maps. Our method achieves performance superior or comparable to LeReS with the median scaling. 
This indicates that our proposed fine-tuning loss works well on not only inverse depth spaces but also depth spaces.               


\begin{table}[t]
    \begin{center}
    \caption{\label{table:LeReS} Experimental results on different pre-trained network}
    \resizebox{0.49\textwidth}{!}
    {
    \begin{tabular}{l|c|cccc|ccc}
    \toprule
    \multirow{2}{*}{Method} & \multirow{2}{*}{Data} & \multicolumn{4}{c|}{Lower is better} & \multicolumn{3}{c}{Higher is better} \\
    & &  Abs Rel & Sq Rel & RMSE & RMSE log & $\delta_{1.25}$ & $\delta_{1.25^2}$ & $\delta_{1.25^3}$ \\     
    \midrule
    LeReS \cite{LeReS} & OL-Market & 0.397 & 0.575 & 1.834 & 0.677 & 0.439 & 0.665 & 0.782 \\
    Ours & OL-Market & \textbf{0.274} & \textbf{0.512} & \textbf{1.395} & \textbf{0.330} & \textbf{0.715} & \textbf{0.897} & \textbf{0.947} \\
    \midrule
    LeReS \cite{LeReS} & OL-Home & 0.473 & 0.697 & 1.550 & 0.675 & 0.416 & 0.619 & 0.744 \\
    Ours & OL-Home & \textbf{0.253} & \textbf{0.116} & \textbf{0.574} & \textbf{0.387} & \textbf{0.550} & \textbf{0.813} & \textbf{0.902} \\
    \midrule
    LeReS \cite{LeReS} & OL-Corridor & 0.308 & 0.164 & 1.461 & 0.595 & 0.481 & 0.727 & 0.825 \\
    Ours & OL-Corridor & \textbf{0.237} & \textbf{0.113} & \textbf{1.106} & \textbf{0.341} & \textbf{0.619} & \textbf{0.850} & \textbf{0.924} \\
    \midrule
    LeReS \cite{LeReS} & OL-Office & 0.224 & 0.114 & 0.527 & 0.513 & \textbf{0.678} & 0.821 & 0.880 \\
    Ours & OL-Office & \textbf{0.216} & \textbf{0.082} & \textbf{0.499} & \textbf{0.373} & 0.633 & \textbf{0.838} & \textbf{0.917} \\
    \bottomrule
    \end{tabular}
    }
    \end{center}
    \centering
    A median scaling is applied for LeReS \cite{LeReS} to choose a scale factor.  
    \vspace{-5mm}
\end{table}

\subsection{Qualitative Results}
We visualize the three results in different experimental settings in Fig. \ref{figure:3D}. To extract scale-aware poses, we use VIO for Fig. \ref{figure:3D}-(a) and (c).   
In Fig. \ref{figure:3D}-(a), our method can generate consistent point clouds over multiple images. This shows that our algorithm can predict metric and dense point clouds along with the whole scene. Moreover, we show the generalizability of our system on diverse scenes collected from the handheld device in Fig. \ref{figure:3D}-(c). We demonstrate the video sequence in a large shopping mall containing stairs and escalators. Thanks to the IMU sensors, our system can provide accurate camera motions in this dataset where visual-based SLAM suffers from the difficulty of predicting reliable motions due to the small field-of-view, rapid camera movements with rotation, and other challenging environments. Also, our depth estimation network can generate metric and coherent depth maps for dense mapping. 

In Fig. \ref{figure:3D}-(b), we show the visualization result of our algorithm on a mobile robot setting. For the experiment, a video sequence is collected with our custom mobile robot equipped with wheel encoders. To estimate accurate robot poses, the pre-integrated encoder is jointly optimized with visual feature reprojection errors in a sliding window-based factor graph optimization scheme \cite{Forster}. The result shows that our algorithm can be compatible with a monocular SLAM system with wheel-odometry measurements regardless of extremely noisy sparse depths. 

\section{Conclusions, Limitations, and Future Work}
We have developed a fine-tuning method for metrically accurate depth estimation in a self-supervised way. Our approach utilizes monocular SLAM with proprioceptive sensors for training and can perform metric depth estimation for dense mapping. The distillation loss enables high-quality depth estimation, and then the photometric and scale-consistent losses are designed for aligning the depth scale with metrically accurate poses. The current work still has limitations to be addressed in future work. Although our knowledge distillation is effective for improving self-supervised methods, it is still based on the abilities of pre-trained depth estimation models. In the future, we will design a back-end multi-view optimization that can overcome the ill-posed problem of a monocular depth estimation and improve the consistency of metrically estimated depths.    


\addtolength{\textheight}{0cm}






\end{document}